\documentclass{article}
\usepackage[ruled,vlined]{algorithm2e}
\usepackage[margin=1in]{geometry}
\usepackage[utf8]{inputenc} 
\usepackage[T1]{fontenc}    
\usepackage{hyperref}       
\usepackage{url}            
\usepackage{booktabs}       
\usepackage{amsfonts}       
\usepackage{nicefrac}       
\usepackage{microtype}      
\usepackage{xcolor}         
\usepackage{graphicx}
\usepackage{amsmath}
\usepackage{xcolor}
\usepackage{wrapfig}
\DeclareMathOperator*{\argmin}{arg\,min}

\begin{document}

Notice: This manuscript has been authored by UT-Battelle, LLC, for the U.S. Department of Energy under contract DE522 AC05-00OR22725. The United States Government retains and the publisher, by accepting the article for publication, acknowledges that the United States Government retains a non-exclusive, paid-up, irrevocable, world-wide license to publish or reproduce the published form of this manuscript, or allow others to do so, for the United States Government purposes. The Department of Energy will provide public access to these results of federally sponsored research in accordance with the DOE Public Access Plan (https://www.energy.gov/doe-public-access-plan).

\title{Active Causal Learning for Decoding Chemical Complexities with Targeted Interventions}

%

\author{%
  Zachary R Fox$^*$ and Ayana Ghosh$^{**}$ \\ 
  Computational Sciences and Engineering Division \\
  Oak Ridge National Laboratory, Oak Ridge, Tennessee, 37831, USA \\
  \underline{Email}: \texttt{$^*$foxzr@ornl.gov}, \texttt{$^{**}$ghosha@ornl.gov}
   \\
}
\date{}

\maketitle

Predicting and enhancing inherent properties based on molecular structures is paramount to design tasks in medicine, materials science, and environmental management. 
Most of the current machine learning and deep learning approaches have become standard for predictions, but they face challenges when applied across different datasets due to reliance on correlations between molecular representation and target properties.
These approaches typically depend on large datasets to capture the diversity within the chemical space, facilitating a more accurate approximation, interpolation, or extrapolation of the chemical behavior of molecules.
In our research, we introduce an active learning approach that discerns underlying cause-effect relationships through strategic sampling with the use of a graph loss function. 
This method identifies the smallest subset of the dataset capable of encoding the most information representative of a much larger chemical space.
The identified causal relations are then leveraged to conduct systematic interventions, optimizing the design task within a chemical space that the models have not encountered previously.
While our implementation focused on the QM9 quantum-chemical dataset for a specific design task—finding molecules with a large dipole moment—our active causal learning approach, driven by intelligent sampling and interventions, holds potential for broader applications in molecular, materials design and discovery.
\section{Introduction}
Over the past few decades, machine learning and deep learning (ML/DL) have played a substantial role in driving advancements in molecular design and discovery.
The expansion of publicly accessible repositories (e.g., PubChem \cite{wang_pubchem_nodate}, ZINC \cite{irwin_zinc20free_2020}, ChEMBL \cite{gaulton_chembl_2012}, QM9 \cite{ruddigkeit_enumeration_2012,hyvarinen2013pairwise} ANI-1x,\cite{smith_ani-1ccx_2020} and QM7-X \cite{hoja_qm7-x_2021}) containing structural and physiochemical data, either computed through quantum mechanical calculations or observed in experiments for thousands to millions of molecules, along with advancements in ML/DL algorithms, has paved the way for modeling a wide spectrum of molecular phenomena.
This includes the representation of molecular interactions, chemical bonding, reaction energy pathways, docking, inverse design of molecules for specific targets, synthesis, and gaining novel insights into mechanisms.
These applications span across diverse fields, encompassing drug discovery \cite{carracedo-reboredo_review_2021,kennedy_application_2008,david_molecular_2020,muratov_qsar_2020} antibiotics \cite{stokes_deep_2020}, catalysts \cite{toyao_machine_2020,yang_machine_2020}, photovoltaics \cite{sun_machine_2019}, organic electronics \cite{Gomez-Bombarelli}, and redox-flow batteries \cite{er_computational_2015}.
The quantitative structure-activity/property relationships (QSAR/QSPR)-type models \cite{muratov_qsar_2020,Sheridan2015,carracedo-reboredo_review_2021} and more recently the generative models have largely contributed to the \textit{in silico} molecular design efforts.
\par
However, most DL/ML methods often heavily rely on \textit{correlations}, which provide statistical associations between variables but do not inherently capture cause-effect relations. 
Understanding cause-effect relationships is crucial for gaining deeper insights into molecular interactions, chemical behaviors, and predicting outcomes accurately in various scientific applications.
In the realm of physical and chemical sciences, integrating \textit{cause-effect relations} with ML/DL workflows is a relatively recent development, with only a handful of recent studies \cite{ghosh_cation_2022,ghosh_natphys_2022,Kalinin_ferro_2022,Kalinin_ferro_domain}.
Incorporating causal models enhances the interpretability and reliability of predictions, especially when extrapolating to datasets beyond the training scope.
Exploring the concept of explainability in ML models involves constructing models that are interpretable by humans \cite{RN371, RN372, RN204, RN369, RN402}.
Causal approaches, which frequently utilize straightforward relationships between variables, inherently offer additional mechanisms to comprehend cause-effect dynamics. 
In the molecular domain, the interpretation of ML models has received minimal attention, underscoring the significance of integrating causal frameworks to enhance both predictability and interpretability \cite{wellawatte2023perspective,wellawatte2022model}.
\par
The importance of exclusive reliance on in-built correlations and integrating explainability is equally relevant in the context of contemporary molecular generative models and others. 
These models have innovatively shifted from conventional string representations of molecules to embedded spaces \cite{david_molecular_2020}, offering comprehensive information about the entire molecular scaffold.
However, the latent embeddings of most generative models are neither smooth nor carry ton of useful information, limiting their utility in direct gradient-based optimization methods for targeted design.
The standard Gaussian processes (GPs) within Bayesian optimization (BO) methods, as combined with generative models for finding optimized solutions, fail to incorporate any prior information of physical or chemical behavior of the system in the process.
Recent work lead by Ghosh at al. \cite{ghosh_discovery_2023} has shown how a physics-augmented GP within a hypothesis-driven active learning workflow can be employed to reconstruct functional behavior over an unknown chemical space (for which prior data may not be available).
Overall, there remains a requirement to integrate explainability and cause-effect relations within these methodologies to comprehensively capture the diversity within chemical space.
This is crucial for achieving a more precise approximation, interpolation, or extrapolation of the chemical behavior of molecules, setting the scope for our current work.
\begin{figure}
    \centering
    \includegraphics[width=.75 \textwidth]{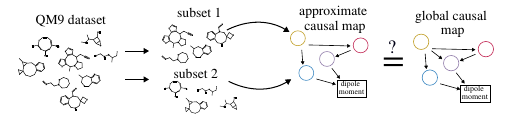}
    \caption{Overview of workflow: Illustration outlining the key steps of the active causal learning approach, which involves constructing a minimal dataset to encapsulate maximal information about molecular structures and properties. This is followed by the active learning of causal relations for the entire dataset, facilitating targeted molecular design.}
    \label{fig:overview}
\end{figure}

\par
In this study, we demonstrate how an active learning workflow, informed with causal discovery models, can successfully learn structure-property relationship from subsets of data, sampled from any part of the chemical space of interest, compare predictive accuracy, all combined together to actively learn the relations for the entire dataset.
We have employed linear causal models to extract causal relations from SMILES and molecular features to perform features selection, extract causal relations for data subsets, link them via active learning with graph metrics, and then finally perform causal interventions for targeted design of molecules.
The target property considered here is dipole moment of a molecule.
For simplicity, we have utilized easily-computable molecular features to represent each molecule.
The choice of data subsets is more or less arbitrary, meaning the information we start with is partial, catering to the adaptive needs for real-time workflows for AI-guided design, automated synthesis, automated characterization while deriving fundamental understandings behind a molecular property.
\section{Results \& Discussion}
The QM9 dataset showcases extensive diversity within chemical space by incorporating a wide range of molecules that encompass different chemical elements and conformational isomers.
It includes information on molecular properties, including dipole moment, derived from quantum-mechanical computations.
As a result, the intricate cause-and-effect relationships between molecular structures, features with properties, exhibit variations across this expansive chemical landscape.
We first demonstrate how the prediction of dipole moment and associated causal relations exhibit such large variability across different regions of the chemical space.
The causal models trained to predict in one region of chemical space do not generalize well to the other regions as represented in Fig.~\ref{fig:generalize}. 
Therefore, causal relations derived from one region may not be robust enough to capture structure-functionality relations for another subset or even the entire dataset.
To address this challenge, we introduce an innovative active learning approach designed to recover comprehensive causal relationships using a minimal dataset, as represented in Fig.~\ref{fig:overview}.

\subsection{Generation of molecular data subsets}
\begin{figure}[ht!]
    \centering
    \includegraphics{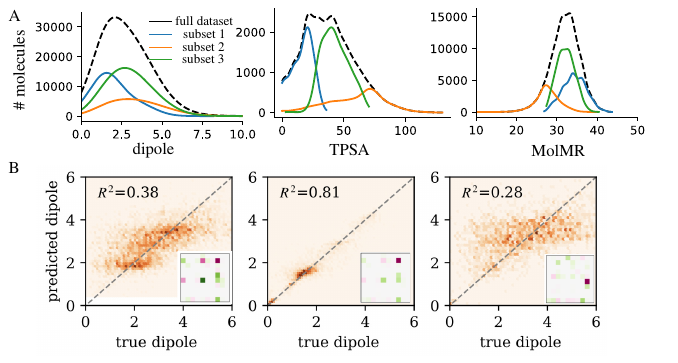}
    \caption{Causal discovery and property prediction in different regions of chemical space. (A) Feature distributions for different subsets. (B) Test $R^2$ and parity plots for a random forest model trained on $\mathcal{D}_1$, $\mathcal{D}_2$, $\mathcal{D}_3$ (left to right). Insets show the corresponding adjacency matrix of the nine downselected features, where green colors signify positive causal relations and pink/purple signify negative relations. Full causal graphs are given in the Appendix. }
    \label{fig:subset_results}
\end{figure}

We begin our analysis by characterizing each molecule within the QM9 dataset \cite{ruddigkeit_enumeration_2012,hyvarinen2013pairwise} through a vector consisting of twenty descriptors, computed using RDKit. 
Instead of employing fingerprint or latent representations of molecules, we choose to work directly with these molecular features, enabling the use of straightforward causal approaches.
Furthermore, different regions of the molecular feature space contribute to the creation of distinct causal maps and predictive models.
We have used a Gaussian Mixture Model to create three subsets, $\mathcal{D}_1, \mathcal{D}_2,\mathcal{D}_3$ of the QM9 dataset by clustering based on three pivotal features: \texttt{MolLogP} (lipophilicity), \texttt{TPSA} (topological polar surface area), and \texttt{MolMR} (molar refractivity) as shown in Fig.~\ref{fig:subset_results}.
After clustering, each data subset retained twenty chemical descriptors. 
\subsection{Feature selection based on predicting polarizability }
We use polarizability as an intermediate target to down-select features for subsequent causal analyses and predictions of dipole moments, thereby capturing the control of an intrinsic property by extrinsic influences.
Using the \texttt{LinGAM} causal discovery framework  \cite{shimizu2006linear}, we pick the top $k \leq 20$ features within the structure equation model\cite{bollen1989structural}.
This approach is valid when each feature has its own additive non-Gaussian noise term $\epsilon_i$, and constructs a model with linear relationships between each variable. 
For each subset of the data $\mathcal{D}_1, \mathcal{D}_2, \mathcal{D}_3$, we have used \texttt{LinGAM} to construct a weighted directed acyclic graph (DAG), denoted by $\mathcal{G}_1, \mathcal{G}_2,\mathcal{G}_3$. In each graph, the target variable (dipole moment) is a sink, i.e. it does not have any downstream variables.
%
%
%
All 20 features are ranked by the strength of their relationships, and the top $k$ are selected.
For the numerical experiments below, we set $k=9$.
This causal analysis is performed for the full dataset, and the same 9 features are used for each data subset.
%

In each data regime, the prediction accuracy of dipole moment using a random forest model over the $k=9$ features is significantly different, as shown in Fig.\ \ref{fig:subset_results}(B). 
Furthermore, we find that each data subset results in distinct causal relationships (Fig.\ \ref{fig:subset_results}B and \ref{fig:subset1}-\ref{fig:subset3}) between the features.
Based on these results, we next investigate how one can construct a minimal dataset that accurately builds causal maps representative of the full dataset.
\subsection{Causally-informed active learning to build minimal molecular datasets}
\begin{figure}[h!]
    \centering
    \includegraphics[width=\textwidth]{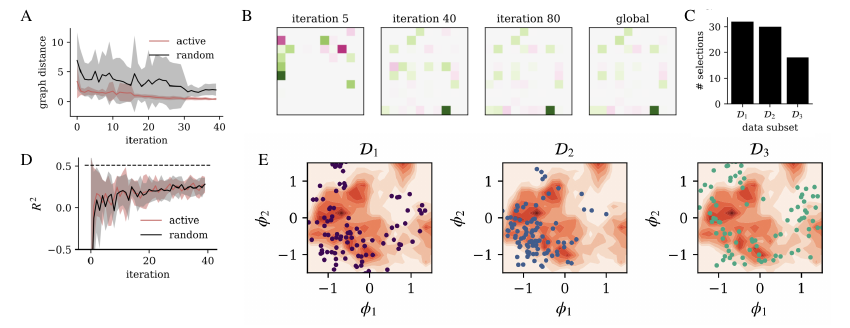}
    \caption{Active learning to recover causal relations. (A) Average and one standard deviation of the graph distance (upper) between the global graph, $\mathcal{G}_\rho$ and the graph of the candidate data set $\mathcal{G}_{\rm AL}$ at each iteration of the active learning algorithm (red) and for randomly selected data (black). (B) Visualization of the adjacency matrices corresponding to $\mathcal{G}_{\rm AL}$ at different iterations and the global DAG $\mathcal{G}_\rho$. (C) The number of times each data subset was selected during the active learning procedure. (D) $R^2$ on test data throughout the active learning experiment. The dashed line corresponds to the value when all data is used. (E) Densities of the ECFPs for the entire dataset projected onto its first two principle components and samples from each data subset (scatter plots). }
    \label{fig:active_learning}
\end{figure}
Active learning aims to optimize the training process by selecting the most informative data points for labeling, rather than relying on random or pre-defined data sampling. In this context, the goal is to reduce the annotation cost and resource requirements while improving model performance. Traditional active learning approaches choose sampling data by evaluating the uncertainty in a predicted value \cite{Lewis1994} or through constructing a dataset which samples the entire input space \cite{ferreira_unsupervised_2016,Ash2019}.  
\par
Here, we build an active learning algorithm, detailed in Algorithm \ref{alg1}, to reconstruct a global causal map from a minimal dataset.
Previous works have emphasized active learning for discovering causal relationships by iteratively performing optimal interventions to distinguish between Markov equivalence classes \cite{Hauser2014,He2008} or even to improve the global conditional structure \cite{Cho2016}. 
%
%
A global causal model may be constructed from existing knowledge about how different molecular features contribute to a target property, such as dipole moment. 
\par
The aim of this active learning algorithm is to reconstruct the global causal map, denoted by $\mathcal{G}_{\rho}$, from a minimal set of data. 
The objective of this algorithm differs from conventional active learning. 
Instead of constructing a minimal dataset solely for predicting a specific property, our aim is to develop a minimal dataset that accurately reproduces causal relationships between structure and property. 
The algorithm uses a graph distance metric, $\mathcal{L}(\mathcal{G}_1,\mathcal{G}_2)$ to compare the global DAG, $\mathcal{G}_{\rho}$ to the DAG $\mathcal{G}_{\rm AL}$ describing the actively learned dataset, $\mathcal{D}_{\rm AL}$.
 $\mathcal{G}_{\rm AL}$ is found using the \texttt{LinGAM} causal discovery framework \cite{shimizu2006linear}.
During each iteration of the active learning scheme, we sample $M$ points, uniformly, from each of the $k$ data subsets described above, denoting the candidate dataset $\tilde{\mathcal{D}}^k_{\rm AL}$. For each candidate dataset, we construct a causal graph, denoted $\tilde{\mathcal{G}}_k $, and compare it to the global graph $\mathcal{G}_{\rho}$ using the graph metric $\mathcal{L}(\mathcal{G}_1,\mathcal{G}_2)$. %
The graph loss function used is the adjacency spectral distance \cite{wills2020metrics}:
\begin{align}
    \mathcal{L}(\mathcal{G}_1,\mathcal{G}_2) = \sqrt{\sum_{i=1}^N \left( \lambda^{\mathcal{A}_1}_i - \lambda^{\mathcal{A}_2}_i \right)^2 },
\end{align}
which computes the $\ell_2$ norm of the top $N$ eigenvalues of adjacency matrix $\mathcal{A}$ for each graph.
\begin{algorithm}[h!] 
\caption{Causally-informed active learning algorithm}
$\mathcal{G}_\rho$, global causal graph \;
$\mathcal{D}_{\rm AL} = \emptyset$ \;
$N_s$, number of data subsets \; 
$N_{\rm iter}$, number of iterations \; 
\While{$n<N_{\rm iter}$}{
    \For{$k \in (1,N_s)$}{
        $\tilde{\mathcal{D}}^k_{\rm AL}= \mathcal{D}_{\rm AL} \cup$ sample($\mathcal{D}_k$,M)\;
        $\tilde{\mathcal{G}}_k = \texttt{LinGAM}(\tilde{\mathcal{D}}_{\rm AL})$ \;
        $s_k = \mathcal{L}(\tilde{\mathcal{G}}_k, \mathcal{G}_\rho) $
    }
    $k^* = \argmin (s_k)$ \;
    ${\mathcal{D}}_{\rm AL} = \tilde{\mathcal{D}}^{k^*}_{\rm AL} $
}
\end{algorithm} \label{alg1}
The dataset with the minimal graph distance to $\mathcal{G}_{\rho}$, denoted by $\tilde{\mathcal{D}}^{*}_{\rm AL}$, is selected, and the algorithm continues. We compare the graph distance of the selected datasets with those chosen from random subsets in Fig.\ \ref{fig:active_learning}A. The actively generated dataset not only converges to the global graph more quickly than the random data (see Fig.\ \ref{fig:active_learning}A-B), but also does so with less noise, as demonstrated by the shaded regions in Fig.\ \ref{fig:active_learning}A. The shaded regions correspond to the mean $\pm$ one standard deviation over ten realizations of the algorithm. Interestingly, the test $R^2$ of the random forest model  performs equally well on either the active or random dataset (Fig.\ \ref{fig:active_learning}(D)), indicating that optimizing for causal structure neither helps nor harms the regression accuracy.
The extended connectivity fingerprints (ECFPs) \cite{rogers_extended-connectivity_2010} over the entire dataset are projected onto their first two principal components, denoted $\phi_1$ and $\phi_2$, to demonstrate that the space explored does not exactly fit into the typical diversity-uncertainty sampling paradigms (Figure \ref{fig:active_learning}C, E).
\begin{figure}[h!]
    \centering
    \includegraphics[width=\textwidth]{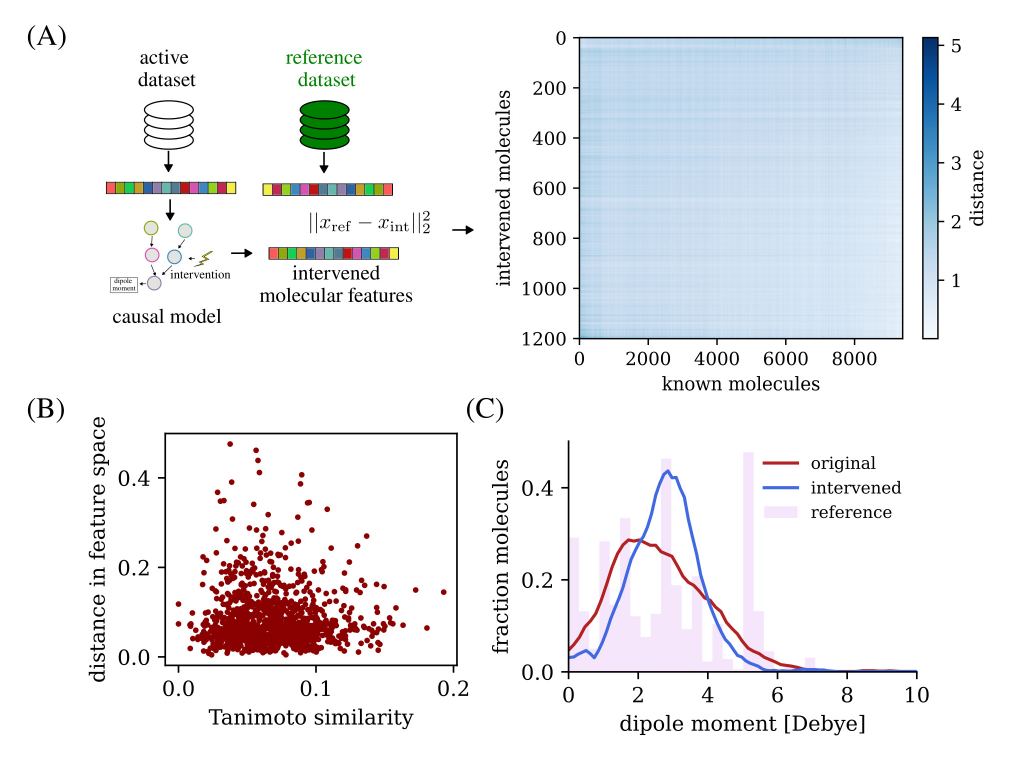}
    \caption{(A) Overview of the method. The $\mathcal{D}_{\rm AL}$ and its associated causal structure is used to find molecular interventions that drive the dipole to the prescribed value. We then search a reference dataset for the molecules which are most similar to the intervened features, shown in the right panel. (B) Scatter plot of the structural similarity between each molecule in $\mathcal{D}_{\rm AL}$ and the reference dataset and the distance in feature space between the intervened molecules and reference molecules. (C) Dipole moments in the original (red) and intervened (blue) datasets. The dipole moments of the closest-to-intervened molecules are show in the pink histogram. }
    \label{fig:dataset_comparison}
\end{figure}
\subsection{Design polar molecules via causal model interventions}
Next, we aim to analyze how the actively learned molecular dataset can be used to design molecules with a high dipole moment, $>$3 Debye. 
Molecules with high dipole moments have potential applications in organic chemistry \cite{minkin2012dipole}, synthesis and drug design \cite{lien1982use} for numerous tasks such as solvent selection, drug-receptor interactions, coatings, adhesives, devices etc.
However, finding molecules with high dipole moments is challenging due to structural-symmetry constraints, required electronegativity balance, potential chemical instability, synthetic challenges, and the availability of suitable building blocks, in addition to achieve a trade-off to keep a high dipole moment while maintaining chemical stability as well as reactivity.
If we can grasp the collective causal influence of various molecular features on the dipole moment and fine-tune them to attain an optimal targeted design, it would help overcome these challenges.
\par
In the context of causal analysis, one can frame this molecular design problem by performing optimal interventions on the causal graph $\mathcal{G}$. An \textit{intervention} is distinct from an \textit{observation} in which an intervention corresponds to fixing a variable $X$, i.e. explicitly leaving the natural data distribution by setting a given variable to a given value. 
The optimal intervention on variable $X_i$ aims to find the value $X_i=x$ for which the target variable $X_j$ achieves a specified value $y$. In our work, we aim to find the optimal intervention for molecules that have a small dipole moment. The variables are the Rdkit-computed molecular features, and the target variable is the dipole moment.   
While the original notions of optimal interventions were based around intervening for the average population response, i.e. $\mathbb{E}[Y | {\rm do}(X) =x)$, here we are concerned with \textit{optimal individual interventions}. 

To this end, we have considered the actively-learned dataset $\mathcal{D}_{\rm AL}$ and found molecular perturbations that increase the dipole moment of a given molecule. 
The molecules are described through the set of features from RDkit as detailed above. 
Utilizing the theory of individual optimal interventions, we find a perturbation for each molecule and feature independently by asking the two following questions to our causal model: 
\begin{itemize}
    \item What feature should be changed for this molecule to have the largest effect on the dipole moment?
    \item To what degree should this feature be changed to induce a desired change in the dipole moment? 
\end{itemize}
\begin{figure}
    \centering
    \includegraphics[width=.7 \textwidth]{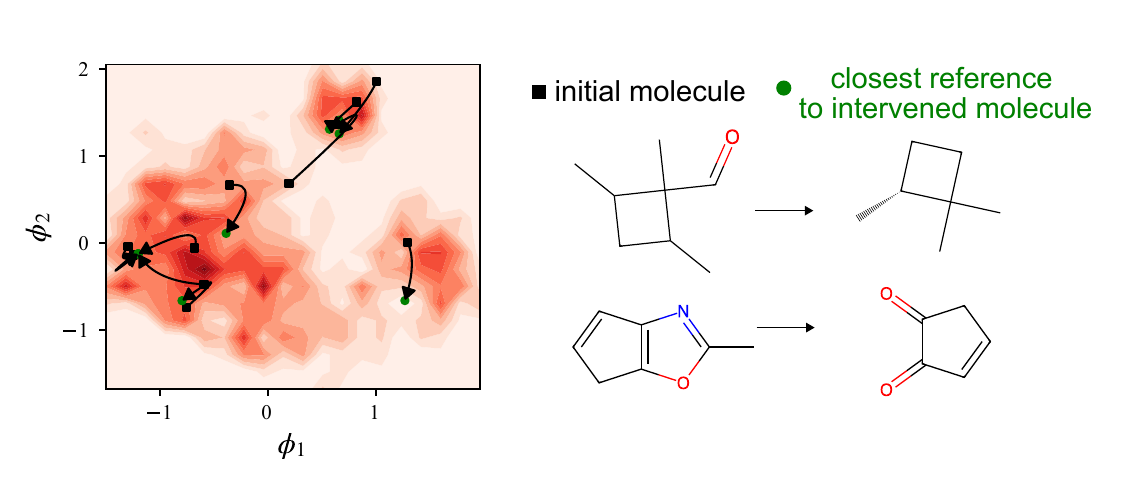}
    \caption{The effect of the interventions in chemical space. The red heatmap is the density of molecules in $\mathcal{D}_{\rm AL}$ projected onto the first two principal components of their ECFPs. Black squares indicate individual molecules from $\mathcal{D}_{\rm AL}$. Green circles indicate the closest reference to the intervened molecules. Two example molecules from $\mathcal{D}_{\rm AL}$ and the closest reference molecules are shown on the right}
    \label{fig:interventions}
\end{figure}
%
After determining what changes must be made to a given molecule (which feature, how much), we apply this change to the features $X_k \rightarrow \tilde{X}_k$. 
However, $\tilde{X}_k$ is not a real molecule; it is the features of a molecule that is predicted to have the desired effect with the prescribed intervention. 
Ideally, one would be able to generate a molecule with the given properties. The  design challenge is to generate or discover such a realistic molecule that captures these properties. This could be done by iterating on the starting molecule \cite{flam2022scalable, chen2021deep, zhu2023pharmacophore,lim2020scaffold}, or by generating a molecule with the desired properties \textit{de novo} \cite{meyers2021novo,weiss2023guided}.

Here, we propose searching over extensive molecular databases for molecules that have similar features vectors to $\tilde{X}_k$. Having identified molecules that have similar feature vectors, we then compare the dipole moments of these molecules to the original molecule to understand if our intervention has yielded the desired effect. 
To this end, we use a subset of 10,000 molecules from the database as a reference dataset, \cite{pereira2018machine} and find the closest molecules to each of the 1,200 perturbed molecules. 
As described in the reference\cite{pereira2018machine}, 
the molecular dipole moment was computed using a semi-automated approach with DFT methods that optimized the 3D structures. The optimized 3D molecular structures were used to compute the necessary partial atomic charges. 
%
%
%

Since $\tilde{X}_k$ is a description of a desirable molecule, rather than the molecule itself, we compare the molecular features with those in the reference dataset using the distance in a normalized feature space. Specifically, we normalize feature $\tilde{X}_k$ across the entire intervened molecular population, and then compute the pairwise distance between the molecular features and each molecule in the reference dataset, return the top $k$ nearest neighbors to each molecule. 

The original molecule (from $\mathcal{D}_{\rm AL}$) and examples of the nearest intervened molecules are shown in Fig.\ \ref{fig:interventions}. 
To understand how similar a given molecule structure is, we have also computed the Tanimoto similarity between the pre-intervention molecules and their downstream "closest" molecules.
Given that the Tanimoto similarity index offers insights into structural similarity, we can further analyze molecular fingerprints by examining their similarity within the space defined by the principal components.
We have calculated $\phi_1$ and $\phi_2$, i.e. the principal components, for the ECFPs of the actively-learned dataset, and show how a given molecule can traverse from its original features space towards the intervened region.
The path from the original molecule to the analog of the intervened molecules are shown in Fig.\ \ref{fig:interventions}.

Finally, we can compare the dipole moments that are predicted for these intervened molecules closest intervened molecules with the dipole moment of the reference molecules that were obtained by DFT computations. 
This evaluation measures the predictive performance of our models in estimating dipole moments based on fundamental molecular features beyond the initial training set.
This methodology prioritizes causal relations over mere correlations, thereby enhancing the model's predictive effectiveness across various datasets.

\subsection{Insights into molecular design}
The actively-learnt causal relations (\ref{fig:subset1}, \ref{fig:subset2}, \ref{fig:subset3}, \ref{fig:subset_all}) also pertain to our chemical understandings. For example, molecules containing NH, OH bonds are highly polar in nature (i.e., a bond dipole) whereas presence of more valence electrons will screen the long-range order, resulting in reduction of dipole moment.
In addition, the presence of complex ring systems or aliphatic chains, hetero atoms may contribute to larger dipole moments due to increased asymmetry and charge distribution.
While molecules with larger differences in electronegativity between atoms tend to have higher dipole moments, the distance between the charge centers are influenced by molecular geometry and the spatial arrangement of atoms.
Hence, tuneable features accounting for molecular weight, atomic charges, number of electrons, electronegativity, presence of NH, OH bonds, collectively contribute towards molecular dipole moment.
We have measured these influences by examining the coefficients of causal relationships between each attribute and the target.
Features (\ref{fig:al_histograms}) with significant cause-effect coefficients have been intervened upon to identify molecules with high dipole moment.
In this context, a dipole moment is deemed significant if it exceeds 3 Debye, a threshold commonly observed in the majority of molecules within the parent QM9 dataset, from which cause-effect relationships have been derived.
We have illustrated a handful of the curated molecules (\ref{fig:interv_mole}) found based on the intervened features with SMILE representations such as
\texttt{N\#CC1=C2C=CC=CN2CCC1=O} ($\sim$10.33 Debye), \texttt{NC1=NC(=S)N[C@H](c2ccccc2)N1} ($\sim$8.76 Debye),
\texttt{CNC1=CN=NC(=O)[C@@H]1Cl} ($\sim$8.53 Debye),
\texttt{O=C1N=C(O)N/C1=C\%C=C\%c1ccccc1} ($\sim$8.38 Debye),
\texttt{O=C(CCl)N1C(=O)N=C(O)[C@H]1O} ($\sim$6.83 Debye),
\texttt{O=S1(=O)N[C@H](c2ccccc2)CCO1} ($\sim$6.43 Debye),
\texttt{C[C@@H]1CN([C@H]2C=C[C@@H](CO)O2)C(=O)NC1=O} ($\sim$5.90 Debye),\\
\texttt{CC1(C)OC(=O)NC[C@H]1c1cccc(O)c1} ($\sim$5.75 Debye),
\texttt{S=P1(NCc2ccncc2)OCCO1} ($\sim$5.47 Debye).
Using the structural features from the 2D SMILE string representations, we can infer valuable insights into the potential (might not be precise) orientation of dipole moments in molecules.
For instance, in \texttt{N\#CC1=C2C=CC=CN2CCC1=O}, the presence of the CN bond suggests a dipole moment directed towards the more electronegative nitrogen atom. Additionally, the overall asymmetry of the molecule due to the arrangement of the aromatic rings and functional groups could influence the direction of the dipole moment.
In \texttt{NC1=NC(=S)N[C@H](c2ccccc2)N1}, it can be influenced by the arrangement of the atoms around the chiral centers in addition to the overall electronegativity differences.
Few of these molecules contain aromatic rings or conjugated double bonds, which may lead to the delocalization of $\pi$-electrons and contribute to distributed dipole moments along the conjugated system. 
Therefore, the directionality of the dipole moment in these molecules may be influenced by the spatial arrangement of the conjugated system within the molecule.

\section{Summary}
In summary, we have developed a causal active learning workflow for iterative identification of causal relations with corresponding prediction of dipole moment for a broad chemical space, from subsets of chemically diverse molecules.
Building upon these robust relationships which go beyond only interpreting the correlations, we systematically intervened on features to pinpoint high-dipole-moment molecules within a distinct dataset, demonstrating the ability of causal models to offer design principles.
It's worth noting that the causal active learning approach isn't restricted to the use of specific features and can be adapted for alternative molecular representations.

This approach brings about two significant advancements in facilitating AI-guided molecular design, synthesis, and characterization.
Firstly, it excels in scenarios where a diverse chemical space is typically utilized to learn about underlying structure-property relationships, which can often be costly and heavily reliant on data fidelity for reasonable predictions.
Secondly, it holds the potential to guide autonomous experiments by adaptively learning causal relations based on partial information from past measurements, aiding in targeted molecular design and synthesis.
Moreover, it proves effective in real-time identification of intrinsic and extrinsic features, allowing scientists to gain insights into the underlying mechanisms governing physical and chemical phenomena.

\section*{Acknowledgements}
This research was sponsored by the SEED (A.G.) and Artificial Intelligence Initiative (Z. R. F.) within Laboratory Directed Research and Development Program of Oak Ridge National Laboratory, managed by UT-Battelle, LLC, for the U.S. Department of Energy. ORNL is managed by UT-Battelle, LLC, for DOE under Contract No. DE-AC05-00OR22725.

\section{Conflicts of interest}
The authors have no conflicts of interest to declare.

\section{Author Contributions}
Z.R.F. and A.G. conceptualized the idea of active causal learning for molecular design. A.G. wrote parts of the preliminary workflow while Z.F. implemented the workflow for the full dataset. Both authors participated in writing the manuscript.

\section{Data Availability}
Data is available at https://zenodo.org/records/10887547
\section{Code Availability}
Code is available at https://code.ornl.gov/z6f/CauMol. 

\newpage
\bibliographystyle{unsrt}
\bibliography{refs}
\appendix\appendix
\renewcommand{\thesection}{S\arabic{section}}
\renewcommand{\thefigure}{S\arabic{figure}}
\setcounter{figure}{0}
\renewcommand{\thetable}{S.\arabic{table}}
\setcounter{table}{0}

\section*{Supplementary Information}

\section{ Structural equation models and causal discovery}
Throughout this work, we use structural equation models of the form 
\begin{align}
    x_i = \sum_{j \in {\rm Pa}(i)} \alpha_j x_j + \epsilon_j,
\end{align}
where $x$ is the observed variable, ${\rm Pa}(i)$ indicates the set of parent indices for the $i^{\rm th}$ variable, and $\alpha_j$ the weight of a given parent's influence. The term $\epsilon_j$ is an additive noise. When all $\epsilon_j$ are independently, non-Gaussian distributed  with non-zero variance, the LinGAM framework for causal discovery may be applied \cite{shimizu2006linear}. The features used in our data set are largely non-Gaussian (see histograms in section xxx).

\newpage
\begin{figure}[h!]
    \centering
    \includegraphics[width=\textwidth]{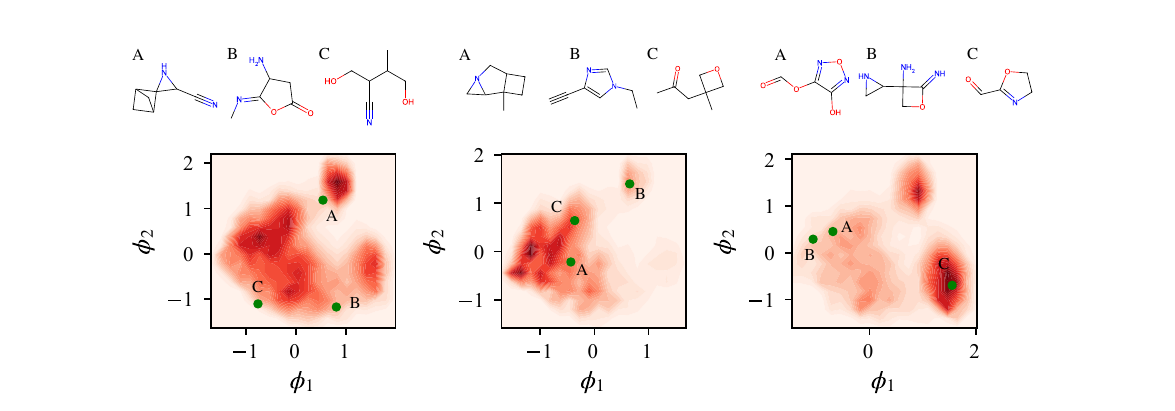}
    \caption{Density plot of the molecules' projected onto the principle components of extended connectivity fingerprint (ECFPs) for each of the data subsets, along with three representative molecules from each data subset.}
    \label{fig:chem_space}
\end{figure}
\newpage
\begin{figure}[h!]
    \centering
    \includegraphics[width=.5\textwidth]{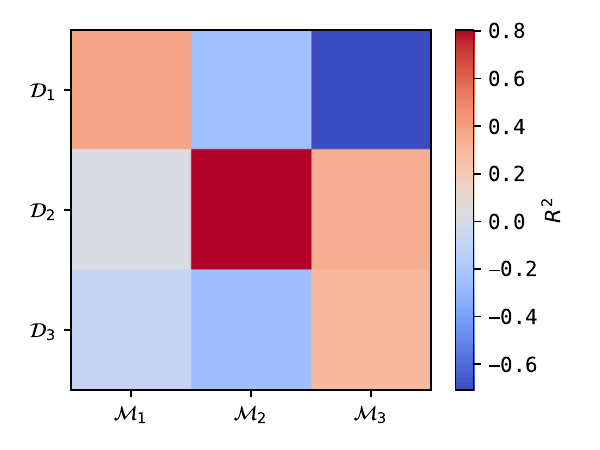}
    \caption{Models trained on different data subsets do not generalize well.}
    \label{fig:generalize}
\end{figure}
\newpage
\begin{figure}[h!]
    \centering
    \includegraphics[width=.8\textwidth]{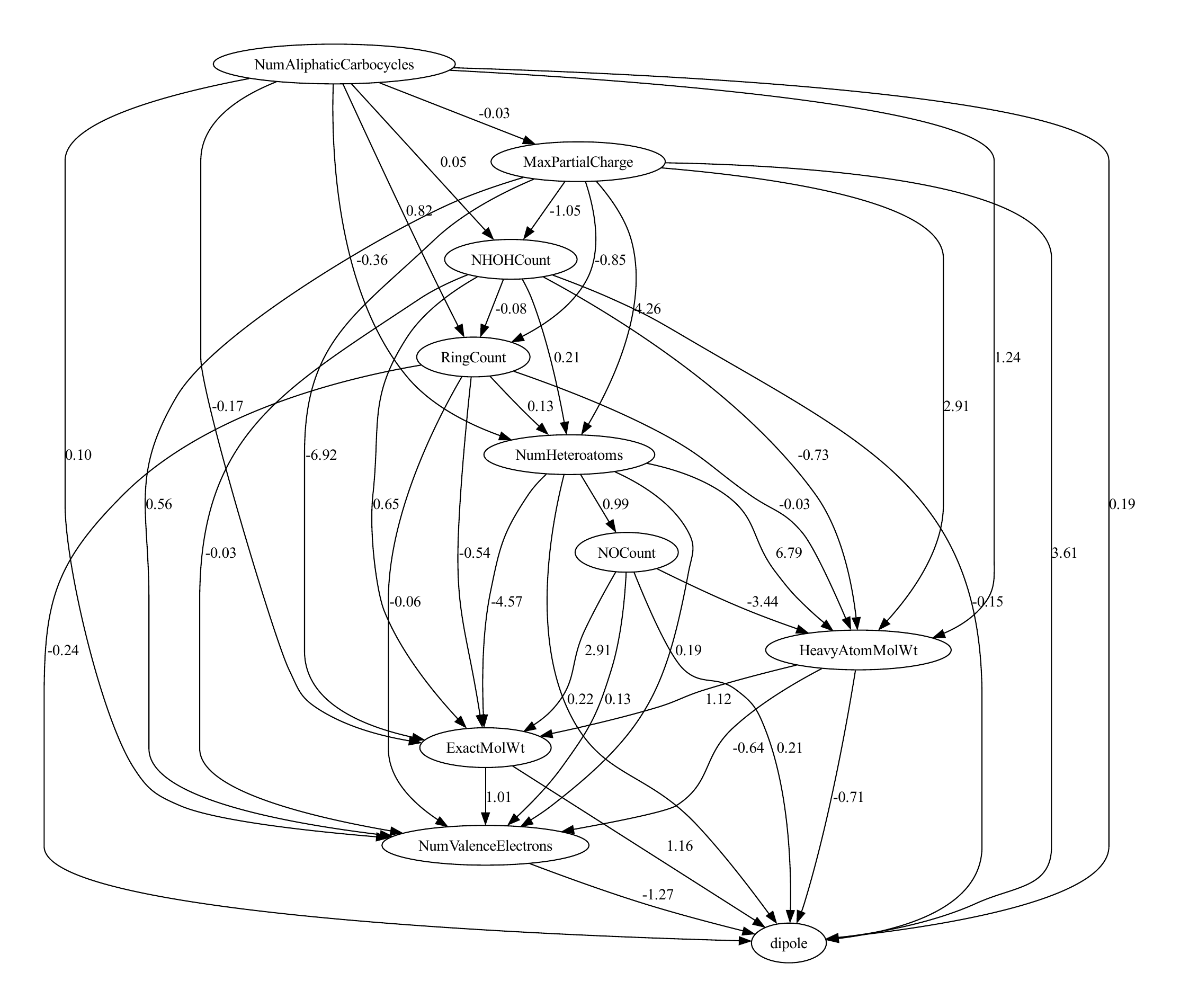}
    \caption{Causal maps associated with $\mathcal{D}_1$.}
    \label{fig:subset1}
\end{figure}
\begin{figure}[h!]
    \centering
    \includegraphics[width=.8\textwidth]{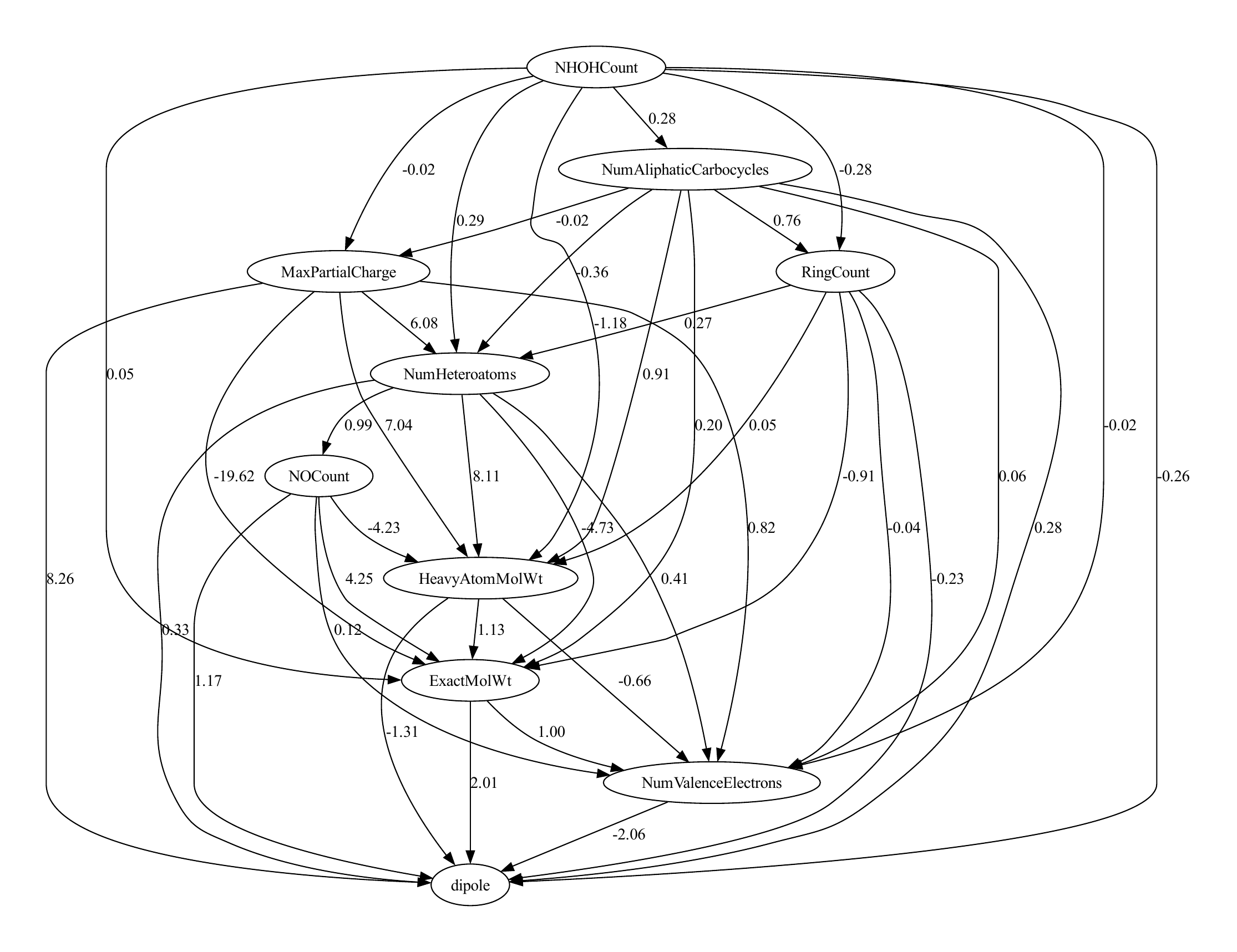}
    \caption{Causal maps associated with $\mathcal{D}_2$.}
    \label{fig:subset2}
\end{figure}
\begin{figure}[h!]
    \centering
    \includegraphics[width=.8\textwidth]{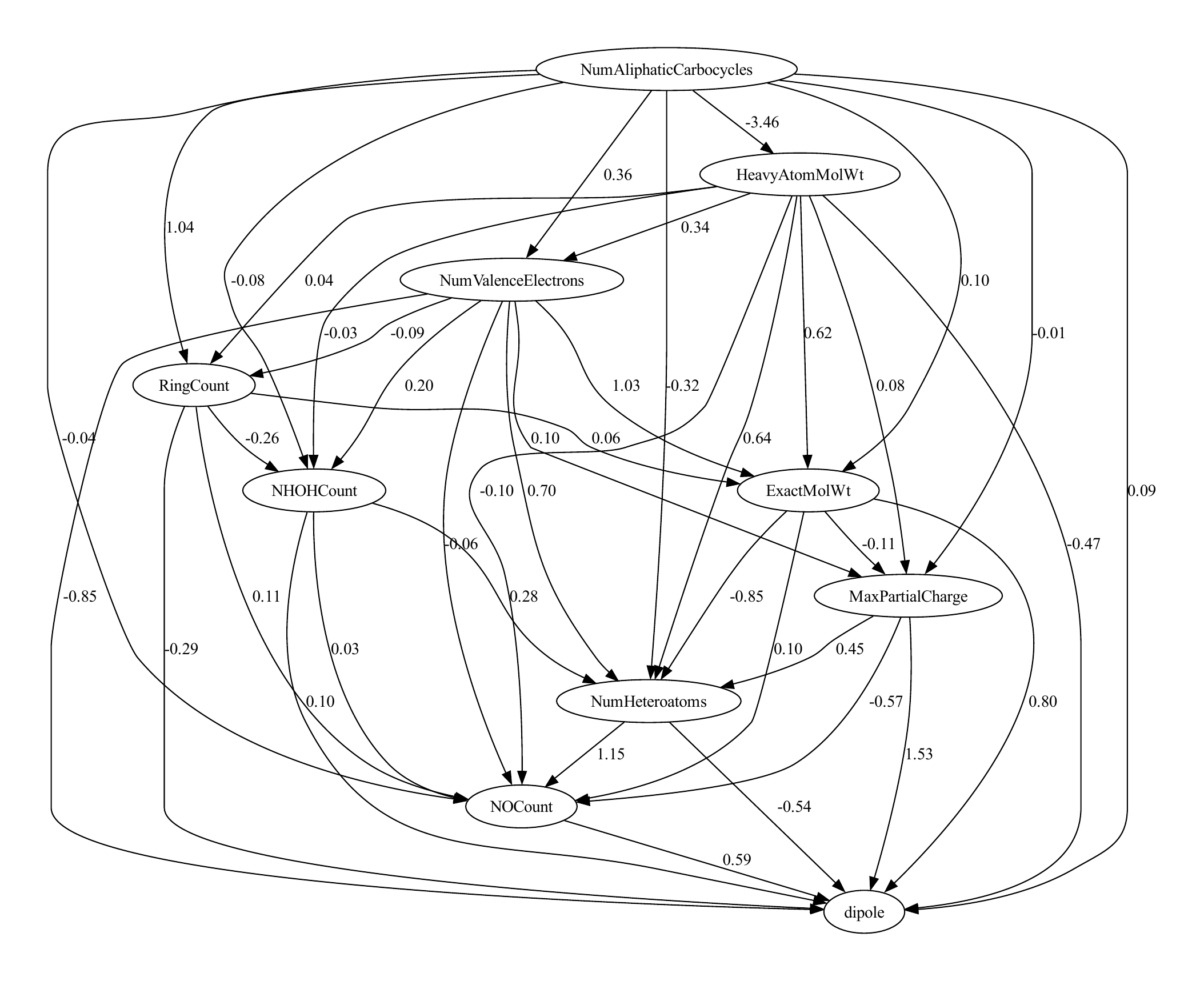}
    \caption{Causal maps associated with $\mathcal{D}_3$.}
    \label{fig:subset3}
\end{figure}
\begin{figure}[h!]
    \centering
    \includegraphics[width=.8\textwidth]{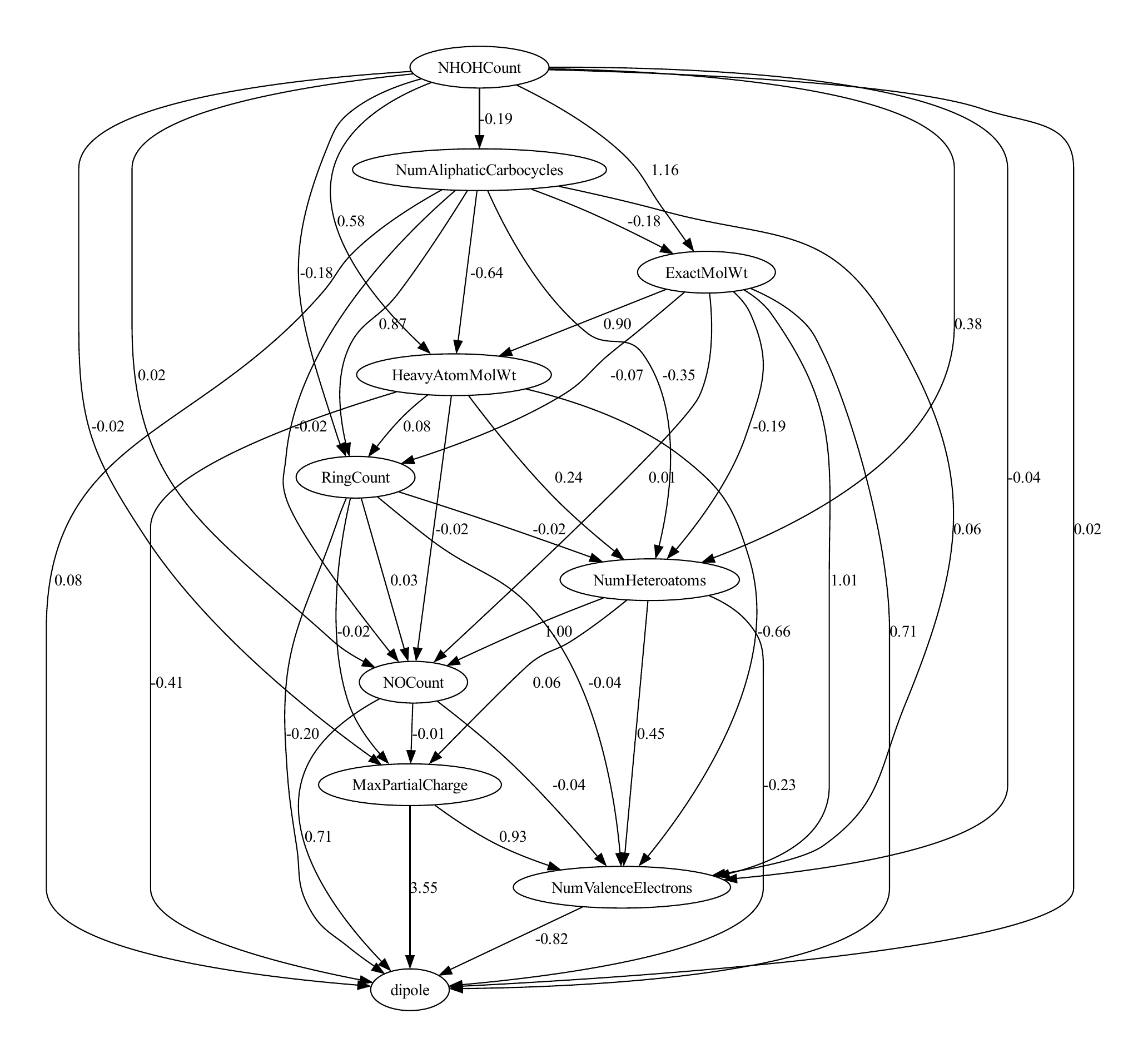}
    \caption{Causal maps associated with $\mathcal{D}_\rho$.}
    \label{fig:subset_all}
\end{figure}
\begin{figure}[h!]
    \centering
    \includegraphics[width=.8\textwidth]{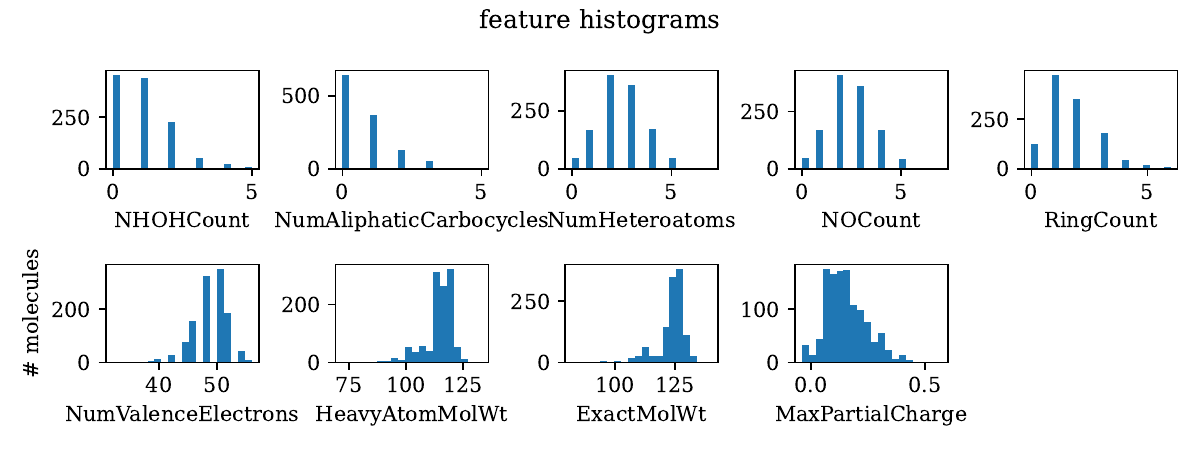}
    \caption{Feature histograms from the actively-learned dataset, $\mathcal{D}_{\rm AL}$.}
    \label{fig:al_histograms}
\end{figure}
\begin{figure}[h!]
    \centering
    \includegraphics[width=.8\textwidth]{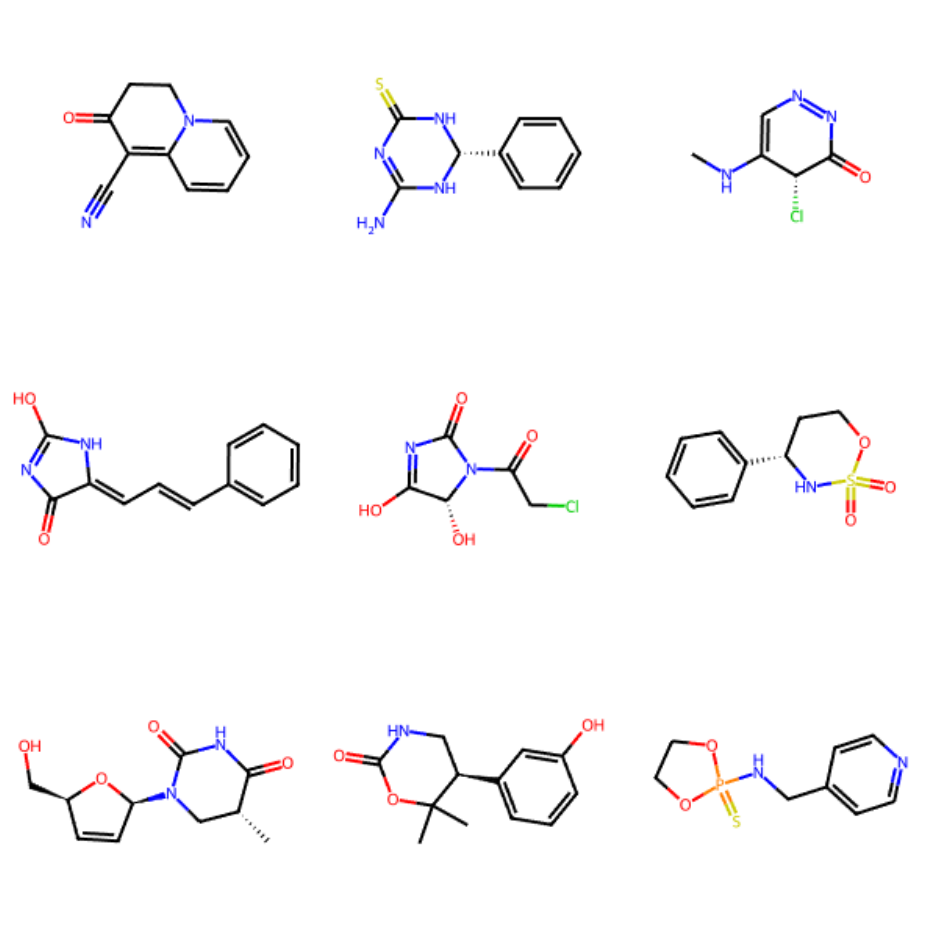}
    \caption{Illustration of a few molecules found after intervention with dipole moment $>$5 Debye.}
    \label{fig:interv_mole}
\end{figure}

\bibliographystyle{apsrev4-1}
\bibliography{refs.bib}
\end{document}